\documentclass[twocolumn,10pt]{article}

\usepackage[utf8]{inputenc}
\usepackage[T1]{fontenc}
\usepackage{lmodern}
\usepackage{geometry}
\usepackage{authblk}
\usepackage[hypertexnames=false]{hyperref}
\usepackage{xurl}
\usepackage{graphicx}
\usepackage{booktabs}
\usepackage{amsmath,amssymb}

\usepackage[numbers,sort&compress]{natbib}
\usepackage{enumitem}
\usepackage{tabularx}
\setlist[itemize]{label={},leftmargin=0pt,itemindent=\parindent,listparindent=\parindent,itemsep=0pt,parsep=0pt,topsep=0pt,partopsep=0pt,align=parleft}
\setlist[enumerate]{label={},leftmargin=0pt,itemindent=\parindent,listparindent=\parindent,itemsep=0pt,parsep=0pt,topsep=0pt,partopsep=0pt,align=parleft}
\usepackage{threeparttable}
\usepackage{eso-pic}
\usepackage{xcolor}
\usepackage{listings}
\usepackage{float}

\definecolor{codegreen}{rgb}{0,0.6,0}
\definecolor{codegray}{rgb}{0.5,0.5,0.5}
\definecolor{codepurple}{rgb}{0.58,0,0.82}
\definecolor{backcolour}{rgb}{0.95,0.95,0.92}
\lstdefinelanguage{json}{
  morestring=[b]",
  showstringspaces=false,
  alsoletter={0123456789.-},
  morecomment=[l]{//},
  morecomment=[s]{/*}{*/},
  morekeywords={true,false,null},
}
\newcommand{\Description}[1]{}

\begin{document}

\AddToShipoutPictureFG*{%
  \AtPageUpperLeft{%
    \raisebox{-2.15cm}[0pt][0pt]{%
      \makebox[\paperwidth]{%
        \hspace*{1.5cm}\includegraphics[height=0.9cm]{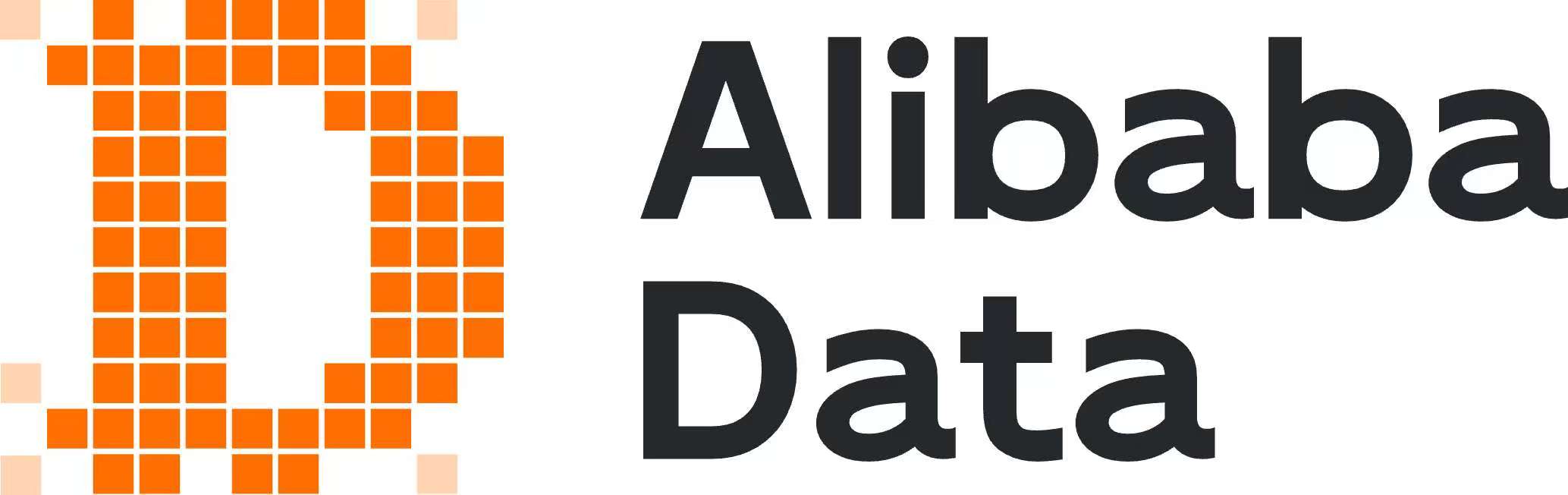}\hfill\includegraphics[height=0.9cm]{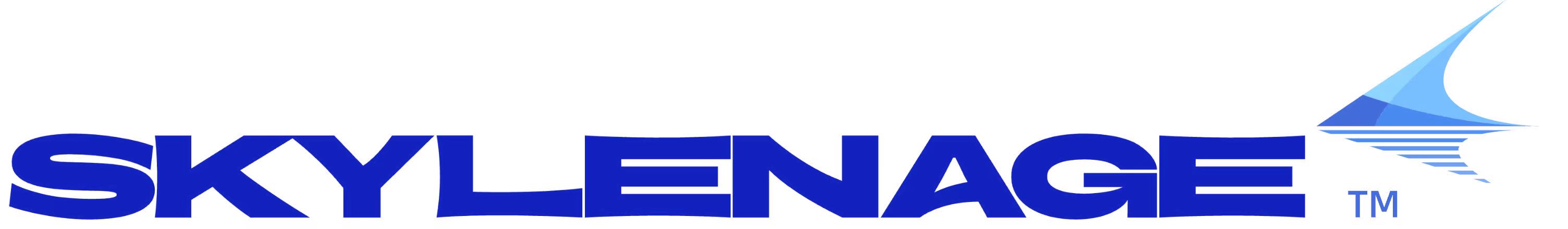}\hspace*{0.8cm}%
      }%
    }%
  }%
}

\title{\textbf{MolQuest: A Benchmark for Agentic Evaluation of Abductive Reasoning in Chemical Structure Elucidation}}

\author[1]{Taolin Han\textsuperscript{*}}
\author[1]{Shuang Wu\textsuperscript{*}}
\author[1]{Jinghang Wang\textsuperscript{*}}
\author[1]{Yuhao Zhou}
\author[1]{Renquan Lv}
\author[1]{Bing Zhao\textsuperscript{\textdagger}}
\author[1]{Wei Hu\textsuperscript{\textdagger}}

\affil[1]{Alibaba Group, Hangzhou, China}
\affil[ ]{{\footnotesize \textsuperscript{*}The first three authors contributed equally to this work.}}
\affil[ ]{{\footnotesize \textsuperscript{\textdagger}Corresponding authors.}}

\date{}

\maketitle

\begin{abstract}
Large language models (LLMs) hold considerable potential for advancing scientific discovery, yet systematic assessment of their dynamic reasoning in real-world research remains limited. Current scientific evaluation benchmarks predominantly rely on static, single-turn Question Answering (QA) formats, which are inadequate for measuring model performance in complex scientific tasks that require multi-step iteration and experimental interaction. To address this gap, we introduce \textbf{MolQuest}, a novel agent-based evaluation framework for molecular structure elucidation built upon authentic chemical experimental data. Unlike existing datasets, \texttt{MolQuest} formalizes molecular structure elucidation as a multi-turn interactive task, requiring models to proactively plan experimental steps, integrate heterogeneous spectral sources (e.g., NMR, MS), and iteratively refine structural hypotheses. This framework systematically evaluates LLMs' abductive reasoning and strategic decision-making abilities within a vast and complex chemical space.Empirical results reveal that contemporary frontier models exhibit significant limitations in authentic scientific scenarios: notably, even state-of-the-art (SOTA) models achieve an accuracy of only approximately 50\%, while the performance of most other models remains below the 30\% threshold. This work provides a reproducible and extensible framework for science-oriented LLM evaluation, our findings highlight the critical gap in current LLMs’ strategic scientific reasoning, setting a clear direction for future research toward AI that can actively participate in the scientific process.
\end{abstract}


\section{Introduction}

Large Language Models (LLMs) are demonstrating significant value in scientific discovery, with their applications emerging as a key frontier in "AI for Science" research~\cite{zheng2025ScientificDiscovery}. Recently, the successive releases of new models with strong reasoning capabilities---such as GPT-5.2~\cite{GPT52}, Gemini 3-Pro~\cite{Gemini3Pro}, and Qwen-3-Max~\cite{qwen3technicalreport}---have made it increasingly critical to explore their practical abilities in complex and dynamic real-world research scenarios.

\begin{figure}[H]
  \centering
  \includegraphics[width=\linewidth]{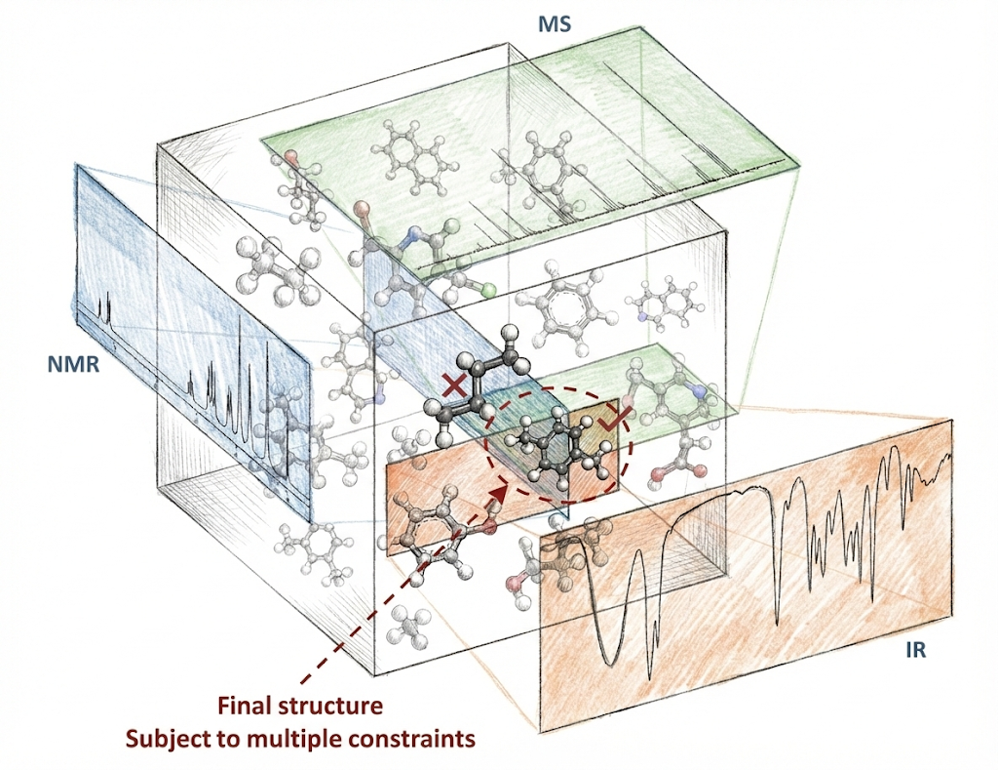}
  \Description{Molecular structure elucidation as a constraint satisfaction problem (CSP), illustrating how multiple experimental constraints jointly narrow down candidate structures.}
  \caption{Molecular structure elucidation as a Constraint Satisfaction Problem (CSP).}
  \label{fig:duoweiyueshutuili}
\end{figure}

A quintessential example of such a core scientific task is molecular structure elucidation in chemistry, which involves deducing the precise structure of an unknown compound from discrete, multimodal experimental data (e.g., mass spectrometry, NMR spectra).~\cite{elyashberg2011contemporary} Molecular structure elucidation is, in essence, a high-dimensional constraint satisfaction problem.~\cite{Lindsay1980Dendral} The model must synergistically integrate experimental data---often incomplete and noisy---from multiple spectroscopic techniques. Together, these data impose a set of chemical constraints that delineate plausible final structures, encompassing key information such as molecular formula, degree of unsaturation, functional group types, and their local chemical environments.~\cite{Pretsch2013Structure} The challenge of this task fundamentally arises from the complex and highly sensitive mapping between chemical structures and spectral signals. A positional isomerism involving an oxygen atom can significantly alter the chemical shifts and coupling constants in the $^{1}$H NMR spectrum; conversely, subtle differences in spectral features may correspond to entirely distinct molecular skeletons~\cite{crews2009organic}. As shown in Figure~\ref{fig:duoweiyueshutuili}, successful structure elucidation therefore requires robust abductive reasoning capabilities: the ability to begin with observed spectral signals, systematically generate chemically plausible candidate structural hypotheses, actively plan and acquire new experimental evidence to introduce additional constraints, iteratively eliminate untenable hypotheses, and ultimately converge on a structure that is logically consistent with all known data under chemical principles.

Although several benchmarks have been proposed to evaluate the scientific capabilities of LLMs, they remain notably limited in assessing a model’s ability to address authentic research problems. We summarize these limitations along three dimensions:

Evaluation format. Mainstream benchmarks (e.g., ChemBench) typically adopt static, single-turn multiple-choice formats, which can lead to difficulty saturation and make performance susceptible to training-data contamination~\cite{mirza2024superhuman}.
Data authenticity. Some benchmarks (e.g., ChemIQ) rely on synthetic simulation data, which fails to capture key real-experiment artifacts such as noise, peak overlap, complex coupling effects, and systematic peak-position deviations~\cite{white2023assessment}.
Lack of initiative. Most existing designs omit the experimental planning and strategic decision-making that are central to real scientific workflows, reducing the model to a passive respondent~\cite{guo2024can}. In practice, scientists operate in a partially observable decision-making environment and must select the most informative next experiment under cost constraints~\cite{song2025evaluatinglargelanguagemodels}.

To address these gaps, we propose MolQuest, an innovative agent-based dynamic benchmark for molecular structure elucidation. Our central aim is to evaluate the abductive reasoning and strategic planning capabilities of ``LLM as a Chemist''---treating the large language model as an autonomous agent capable of operating within a simulated laboratory environment, thereby examining its scientific problem-solving abilities in dynamic real-world research contexts.

\begin{figure}[H]
  \centering
  \includegraphics[width=\linewidth]{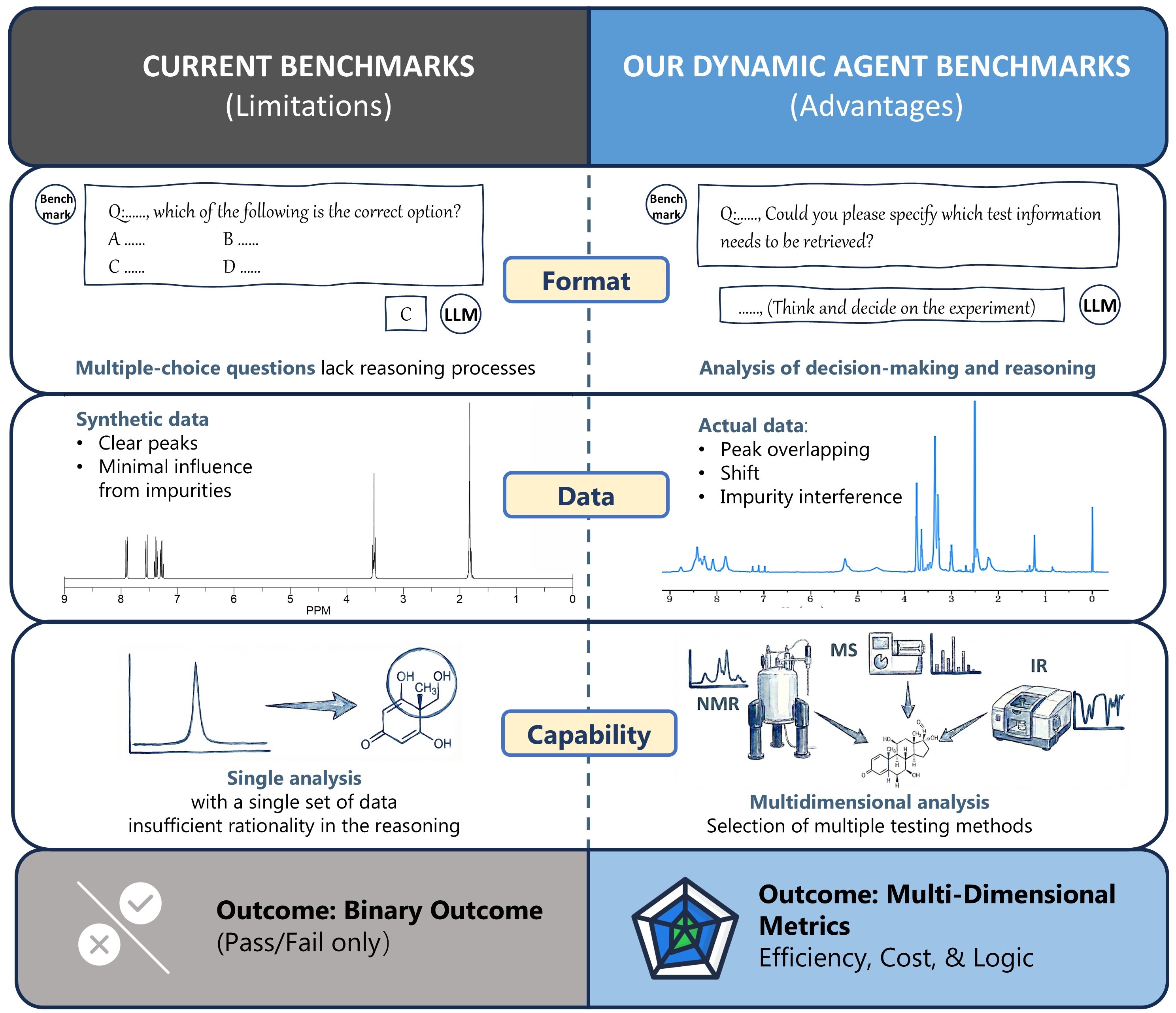}
  \Description{Overview of the MolQuest benchmark characteristics and key components.}

  \caption{The characteristics of our benchmark.}
  \label{fig:benchmark-characteristics}
\end{figure}

Unlike existing benchmarks, MolQuest is characterized by three core features, as Figure~\ref{fig:benchmark-characteristics}:

Dynamic Interactive Paradigm: The model operates as an autonomous agent within a virtual laboratory, actively invoking tools (e.g., ``Measure Molecular Weight,'' ``Obtain $^{1}$H NMR Spectrum'') to acquire information as needed while iteratively refining its structural hypotheses in the process.

Real-Data-Driven Construction: To ensure both realism and fairness in evaluation scenarios, we established a rigorous human-in-the-loop data pipeline. Over half of all test cases are extracted and validated from supporting information in chemical literature published after 2025, effectively mitigating the risk of data contamination.

Multi-Dimensional Capability Assessment: We conducted a comprehensive evaluation of 12 state-of-the-art LLMs on MolQuest. Moving beyond simple final-answer accuracy, the evaluation framework incorporates multiple metrics---including evaluated the decision-making logic and reasoning capabilities of LLMs in complex environments.

\section{Related Work}

\subsection{Contemporary Large Reasoning Models}

The field of artificial intelligence is undergoing a paradigm shift toward reasoning-centric architectures. The new generation of large-scale language models dedicates substantially increased computational resources during the inference phase to tackle complex logical challenges. Leading models such as the OpenAI o-series and GPT-5.2~\cite{GPT52} have moved beyond traditional pattern matching, adopting inference-time strategies like chain-of-thought and search to enable structured problem decomposition. Gemini 3 Pro\cite{Gemini3Pro} maintains logical coherence in multi-step tasks through a fine-grained thinking mechanism with adjustable depth. Claude 4.5~\cite{Claude45} Opus employs a hybrid reasoning architecture where its deliberate approach strengthens logical verification while preserving practicality. Open-source models have also achieved significant breakthroughs: DeepSeek-R1~\cite{Guo_2025} utilizes reinforcement learning to encourage self-correction during generation, while models like Qwen3-Max~\cite{qwen3technicalreport} dedicated symbolic reasoning modules, substantially enhancing their capacity to handle complex scientific problems. Collectively, these advances represent a transition from passive content generation to active problem-solving, establishing a more robust computational foundation for scientific discovery.

\subsection{Evolution of General Reasoning Benchmarks}

The evaluation framework for large language models is systematically evolving from knowledge retrieval to deep reasoning assessment. Early benchmarks such as GPQA test deep comprehension through "search-proof" questions requiring doctoral-level expertise.~\cite{rein2023gpqagraduatelevelgoogleproofqa} Extending evaluation across dozens of disciplines, Humanity's Last Exam reveals persistent gaps in models' academic capabilities even with multimodal inputs.~\cite{phan2025humanitysexam} Meanwhile, ARC-AGI-2 exposes structural deficiencies in the fundamental reasoning of current AI systems by testing compositional generalization in knowledge-free environments.~\cite{chollet2026arcagi2newchallengefrontier} This evolutionary trajectory clearly indicates that the focus of assessment has shifted from surface-level knowledge coverage to the systematic measurement of deep reasoning quality.

\subsection{Evaluations in the Chemical Domain}

Within the AI for Science framework, the assessment of chemical intelligence has evolved significantly from basic capability testing to complex problem-solving. General chemistry benchmarks like ChemBench~\cite{mirza2024superhuman} and ChemEval ~\cite{huang2024chemevalcomprehensivemultilevelchemical} foundational evaluation systems covering broad curricular content, while specialized frameworks such as QCBench~\cite{Xie_2025} and ChemLLMBench~\cite{guo2023largelanguagemodelschemistry} focus more on quantitative computation and property prediction tasks. To deeply evaluate molecular understanding, ChemIQ~\cite{white2023assessment} and FGBench~\cite{liu2025fgbenchdatasetbenchmarkmolecular} are specifically designed for functional group identification, whereas MolPuzzle~\cite{guo2024can} simulate real structure elucidation processes and ChemCoTBench~\cite{li2026chemicalqaevaluatingllms} simulate multi-step reasoning tasks. Current chemical domain evaluations still face two key limitations: on one hand, even benchmarks emphasizing abductive reasoning like NMR-Challenge~\cite{doi:10.1021/acs.jchemed.4c00092} are constrained by the use of synthetic data; on the other hand, existing agent-based systems such as CHEMAGENT~\cite{tang2025chemagentselfupdatinglibrarylarge} primarily focus on forward prediction in idealized environments, failing to adequately capture the iterative decision-making inherent in authentic scientific research. While MaCBench~\cite{Alampara2025Probing} provides a comprehensive evaluation of experimental workflows, it lacks a metric for measuring model selection strategies.

\section{MolQuest Framework}

\subsection{Benchmark Overview}
Existing evaluations of Large Language Models (LLMs) in scientific discovery predominantly rely on static, single-turn question-answering (QA) formats. However, such benchmarks are inadequate for assessing the proactive planning, strategic decision-making, and iterative reasoning capabilities essential for authentic, dynamic scientific research.

To bridge this gap, we introduce \textbf{MolQuest}, a novel benchmark designed to evaluate the cognitive capabilities of the ``LLM-as-a-Chemist.'' Our core philosophy reframes the classic problem of molecular structure elucidation from a static QA task into a sequential decision-making process. By constraining the agent with real experimental data and costs, MolQuest establishes a dynamic evaluation paradigm that mirrors actual laboratory workflows.

The benchmark's complexity stems from its integration of authentic, multi-modal experimental data—including raw NMR and MS spectra alongside their textual interpretations. As illustrated in Figure~\ref{benchmarkgouzaoguochen}, these data are modularized to serve as environmental feedback, creating reasoning challenges that closely approximate the noise and ambiguity of real-world scenarios.

Unlike traditional one-shot evaluations that provide all evidence upfront, MolQuest immerses the model in a simulated research environment characterized by information asymmetry. Key spectral data are not disclosed initially; instead, they must be requested on-demand based on the agent's evolving hypotheses. This design compels the agent to perform abductive reasoning under conditions of incomplete information and resource constraints (simulating experimental costs), akin to a human chemist.

To achieve this, the construction of MolQuest rests on two complementary pillars:

1. Authentic, Traceable Scenarios: Moving beyond synthetic examples, we construct evaluation tasks directly from the Supporting Information of high-quality chemical literature (detailed in Section 3.2). This ensures that the inherent characteristics of real research—such as data noise, spectral overlap, and information gaps—are faithfully preserved.

2. A Cognitive-Simulation Framework: We design a state-machine-driven environment (detailed in Section 3.3) that enforces a ``Plan--Request--Reason'' loop. This framework shapes the agent's behavior through specific instructions and simulated tools, directly evaluating its ability to translate static knowledge into dynamic, cost-aware problem-solving.

In summary, by integrating complex real-world data with an interactive decision-making mechanism, MolQuest establishes a rigorous testbed for scientific AI that evaluates whether large language models can transcend passive knowledge retrieval to exhibit the proactive, strategic, and resource-efficient reasoning required for autonomous scientific discovery.

\subsection{Task Definition and Data Construction}

\begin{figure}[H]
  \centering
  \includegraphics[width=\linewidth]{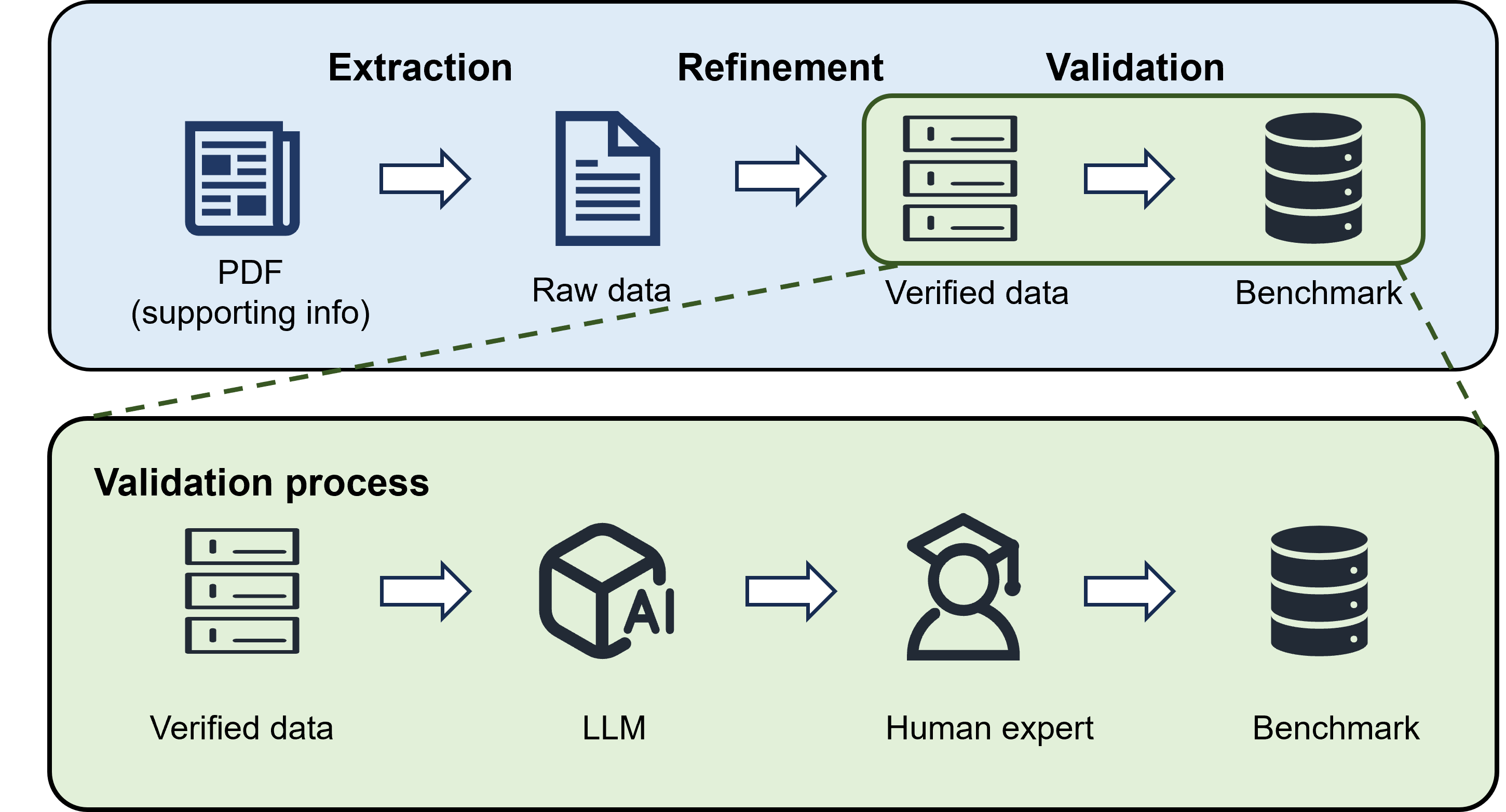}
  \Description{Pipeline diagram showing the construction process of MolQuest from literature sources to validated cases.}
  \caption{Data processing pipeline}
  \label{benchmarkgouzaoguochen}
\end{figure}

In this work, we formalize molecular structure elucidation as a \textbf{Constraint Satisfaction Problem (CSP)}. Given an initial unknown sample, the agent's objective is to identify a chemically valid molecular structure that is consistent with a series of spectral evidence (e.g., NMR, MS) obtained through simulated experiments.

During the data extraction process, the Large Language Model (LLM) outputs exhibited several systematic errors that necessitated manual human intervention for correction.
Misinterpretation of exchangeable protons. The model frequently failed to identify missing hydroxyl ($-\mathrm{OH}$) or carboxylic acid ($-\mathrm{COOH}$) protons within NMR datasets.
SMILES atom counting errors. Significant enumeration errors occurred during atom counting when the model processed SMILES strings.
Peak deviation false alarms. In spectroscopic analysis, the model often incorrectly flagged valid peaks as erroneous due to minor but acceptable chemical shift deviations.
Mass spectrometry adduct confusion. The model consistently confounded the absolute molecular mass ($M$) with observed adduct ion peaks, such as $[M+\mathrm{H}]^+$ or $[M+\mathrm{Na}]^+$.
These failure modes underscore the necessity for rigorous human oversight when utilizing LLMs to handle complex chemical informatics.

To ensure high fidelity and scientific accuracy, we constructed a library of evaluation scenarios derived from recent chemical literature. We implemented a rigorous \textbf{Human-in-the-Loop data pipeline} (Figure~\ref{benchmarkgouzaoguochen}) that combines the efficiency of LLM-based automation with the precision of expert validation. This pipeline consists of three distinct phases:

Phase 1: Automated Extraction and Structuring.
We employed a multi-agent LLM system to parse raw Supporting Information (SI) PDFs. A collaborative architecture—comprising a \textit{Segmenter} to isolate individual molecular entries, a \textit{Spectroscopist} to extract IUPAC names and spectral data, and a \textit{Judge} for real-time error checking—transforms unstructured text into structured JSON records. This automated stage handles the high-volume ingestion of raw data.

Phase 2: Chemical Intelligence and Verification.
Extracted data undergo strict validation using cheminformatics tools. IUPAC names are converted to SMILES via authoritative APIs (PubChemPy) or rule-based parsers (OPSIN), explicitly prohibiting LLM hallucination. Theoretical properties (e.g., MW, Formula) are computed via RDKit as ground truth. Subsequently, an auxiliary LLM performs logical consistency checks—such as verifying proton counts in $^{1}$H NMR against the molecular formula—to flag potential contradictions.

Phase 3: Human-in-the-Loop Final Review.
Given the complexity of spectral interpretation, automated validation cannot guarantee 100\% accuracy. Therefore, all system-flagged entries ("red-flag" data) are routed to a visual review interface. Human chemistry experts examine the original text and extracted structures to correct subtle errors (e.g., stereochemistry or peak assignment ambiguities), serving as the ultimate safeguard for benchmark quality.

Through this pipeline, we established a robust dataset of 530 validated molecular elucidation tasks, covering a diverse chemical space with molecular weights ranging from 150 to 500 Da. This library provides a reliable foundation for evaluating dynamic scientific reasoning.

\begin{figure}[H]
  \centering
  \includegraphics[width=\linewidth]{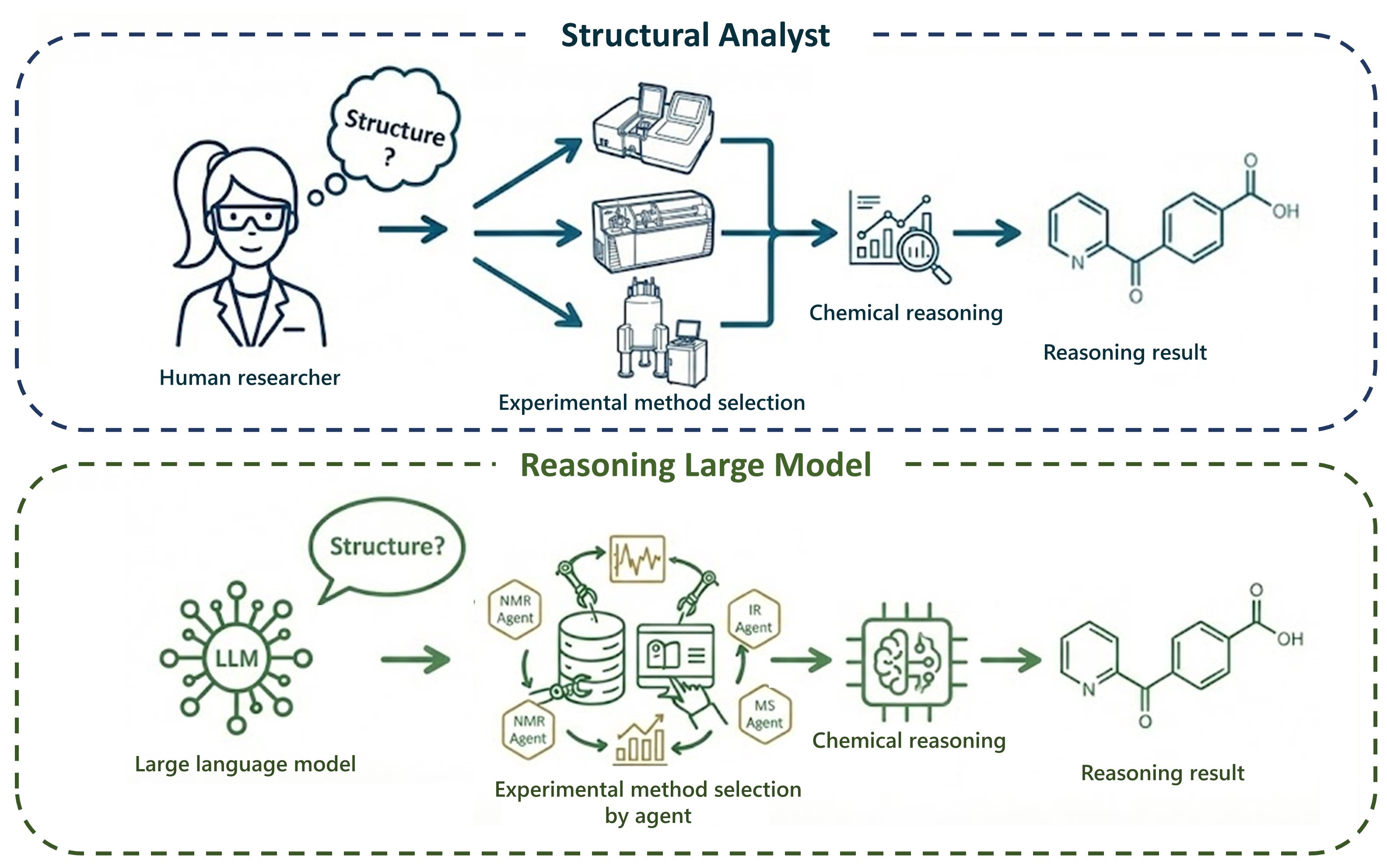}
  \Description{Molecular structure elucidation as a constraint satisfaction problem (CSP), illustrating how multiple experimental constraints jointly narrow down candidate structures.}
  \caption{Comparison of Chemical Reasoning and Decision-Making Between Large Language Models and Human Chemists in Molquest}
  \label{fig:agent}
\end{figure}

\subsection{Interactive Agent Framework Design}
\label{sec:3.3}
To assess dynamic reasoning, we construct an interactive simulation environment modeled as a state machine. In this setup, the LLM acts as a ``Senior Spectroscopist'' tasked with determining molecular structures under resource constraints.

\subsubsection{Agent Persona and Objectives}
Via a detailed system prompt (Appendix C.3), the agent is instructed to operate within a simulated laboratory. Its behavior follows three core principles:

Active Planning: Rather than receiving all data at once, the agent must strategically query specific experimental tools based on current uncertainty. The complete action space consists of 14 simulated instruments, as detailed in Table~\ref{tab:tool_library}.
Iterative Abduction: The agent adheres to a ``Hypothesize--Validate--Refine'' loop.~\cite{Wang2023Scientific} It continuously integrates new evidence to update its structural hypotheses and confidence levels.
Termination and Output: The agent autonomously decides when sufficient evidence is gathered. It then terminates the task by submitting a structured \texttt{FINAL\_RESULT} containing the predicted SMILES string and a self-assessed confidence score (0--100\%).

\begin{table}[t]
  \centering
  \caption{The Action Space: List of simulated experimental tools available to the agent in MolQuest.}
  \label{tab:tool_library}
  \small
  \begin{tabularx}{\linewidth}{@{}lX@{}}
    \toprule
    \textbf{Tool Name} & \textbf{Description} \\
    \midrule
    \texttt{Check\_Data} & Check available data types in the database for the current molecule. \\
    \texttt{Measure\_MW} & Measure molecular weight (simulating Mass Spectrometry). \\
    \texttt{Measure\_Formula} & Measure molecular formula (simulating High-Resolution MS). \\
    \texttt{Calculate\_DBE} & Calculate the Degree of Unsaturation (DBE). \\
    \midrule
    \texttt{Get\_1H\_NMR} & Acquire ${}^1$H NMR spectrum data (proton signals). \\
    \texttt{Get\_13C\_NMR} & Acquire ${}^{13}$C NMR spectrum data (carbon signals). \\
    \texttt{Get\_19F\_NMR} & Acquire ${}^{19}$F NMR spectrum data (fluorine signals). \\
    \texttt{Get\_31P\_NMR} & Acquire ${}^{31}$P NMR spectrum data (phosphorus signals). \\
    \texttt{Get\_IR} & Acquire Infrared Spectroscopy (IR) data. \\
    \midrule
    \texttt{Get\_HRMS} & Acquire full High-Resolution Mass Spectrometry data. \\
    \texttt{Get\_MS} & Acquire standard Mass Spectrometry data. \\
    \midrule
    \texttt{Get\_Melting\_Point} & Acquire melting point data. \\
    \texttt{Get\_TLC} & Acquire Thin-Layer Chromatography (TLC) data. \\
    \texttt{Get\_Optical\_Rotation} & Acquire optical rotation data. \\
    \bottomrule
  \end{tabularx}
\end{table}

\subsubsection{Interaction Mechanism}
The interaction is formalized as a sequential decision process. At each step~$t$, the state~$S_t$ consists of the dialogue history and currently acquired data. The agent observes~$S_t$ and executes an action~$a_t$ (either a tool call from Table~\ref{tab:tool_library} or a final answer). If a tool is called, the environment retrieves the corresponding real-world spectral data, updates the state to~$S_{t+1}$, and incurs a simulated cost. This cycle continues until termination, directly measuring the model's ability to translate static knowledge into dynamic problem-solving.~\cite{Bran2024Augmenting}

\subsection{Evaluation Protocol and Metrics}
To provide a comprehensive assessment of the ``LLM-as-a-Chemist,'' we define a set of metrics covering structural correctness, chemical plausibility, and probabilistic reliability, as reported in our main experiments.

\subsubsection{Chemical Correctness and Plausibility}

Structure Accuracy: The primary success metric, defined as the percentage of cases where the predicted SMILES string is canonically identical to the ground truth.~\cite{smile}
SMILES Validity Rate: The proportion of generated SMILES strings that can be successfully parsed into valid chemical graphs by cheminformatics toolkits (e.g., RDKit), indicating the model's grasp of chemical syntax.
Formula Conservation: A logical consistency metric measuring the percentage of predictions where the molecular formula of the generated structure exactly matches the ground truth formula. This serves as a critical check for hallucination, ensuring the model respects the mass-balance constraints imposed by experimental data.
Average Similarity: For incorrect predictions, we calculate the Tanimoto similarity~\cite{Bajusz2015Why} (based on Morgan fingerprints) between the predicted and ground truth structures. This metric quantifies partial success, indicating how chemically close the model's hypothesis was to the correct answer.

\subsubsection{Probabilistic Reliability}
Since autonomous agents must gauge their own certainty, we evaluate the reliability of their self-assessment using Root Mean Square Calibration Error (RMSCE)~\cite{Guo2017Calibration}. This metric assesses the alignment between the agent's reported confidence scores and its actual accuracy. A lower RMSCE indicates a better-calibrated model that assigns high confidence to correct answers and low confidence to incorrect ones---a trait essential for reliable scientific automation.

\section{Experiments}

This section evaluates state-of-the-art LLMs on MolQuest to assess their absolute capabilities, reliability, and the specific impact of the dynamic agentic paradigm compared to static baselines.

\subsection{Experimental Setup}

\subsubsection{Dataset}
The MolQuest benchmark used in this study consists of 530 independent molecular elucidation cases extracted from the Supporting Information of recent (post-2025) high-quality chemical literature (see Appendix A). It is intended solely for evaluation, with no train/val split. The dataset covers a molecular weight range of 150-500 Da and includes diverse functional groups (e.g., carbonyls, hydroxyls, aromatic rings, nitrogen-containing heterocycles) and chiral centers, ensuring broad coverage and complexity in chemical space (see Figures 1-3 in the Appendix A for chemical space visualizations).
\subsubsection{Evaluated Models}
We evaluated twelve state-of-the-art general-purpose LLMs of varying scales and families: Claude Opus 4.5~\cite{ClaudeOpus45}, Gemini 3 Pro~\cite{Gemini3Pro}
, Claude Sonnet 4.5~\cite{ClaudeSonnet45}, Gemini 3 Flash~\cite{Google3Flash}, Claude Haiku 4.5~\cite{ClaudeHaiku45}, DeepSeek V3.2~\cite{DeepSeekV32}
, DeepSeek V3.1~\cite{DeepSeekV31}
, Qwen3 Max~\cite{qwen3technicalreport}
, Gemini 2.5 Pro~\cite{Gemini25Pro}
, Kimi K2 Thinking~\cite{KimiK2Thinking}
, DeepSeek V3.2 Thinking~\cite{DeepSeekV32},GPT-5.2~\cite{GPT52}.
The temperature was set to 0 for all models to increase determinism.

\subsubsection{Evaluation Configurations}
We conduct evaluations under two distinct configurations:

Agent (Dynamic Interactive): The primary setup where models operate within the \textit{MolQuest} interactive framework (Section~\ref{sec:3.3}). No hard limit is placed on interaction rounds; the agent decides when to terminate by submitting a \texttt{FINAL\_RESULT}. This jointly assesses final correctness, decision confidence, and interaction efficiency.
Baseline (Static One-shot): An ablation setup where models receive \textit{all} relevant spectral data at once and are prompted to output the structure directly (prompt in Appendix C.4). This serves as a control to isolate the effect of dynamic interaction.

Metrics: We report Structure Accuracy (Exact SMILES Match), Validity Rate, Average Similarity (Tanimoto), Calibration Error, and Formula Conservation (consistency between predicted structure and ground truth formula).

Ablation Rationale: The comparison between \textit{Agent} and \textit{Baseline} is designed to decouple foundational chemical knowledge from strategic reasoning. While the Baseline measures the model's ability to map a complete set of spectroscopic data to a structure (pattern matching), the Agent configuration evaluates the model's capacity for hypothesis-driven information acquisition and sequential logic.

\begin{table*}[t]
  \centering
  \caption{Main results on MolQuest under the Agent and Baseline settings.}
  \label{tab:main-results}
  \begin{threeparttable}
  {\scriptsize
  \setlength{\tabcolsep}{3pt}
  \resizebox{\textwidth}{!}{%
  \begin{tabular}{lcccccccccc}
    \toprule
    \textbf{Model} & \multicolumn{2}{c}{\textbf{Accuracy (\%)}} & \multicolumn{2}{c}{\textbf{Validity Rate (\%)}} & \multicolumn{2}{c}{\textbf{Average Similarity (\%)}} & \multicolumn{2}{c}{\textbf{Calibration Error (\%)}} & \multicolumn{2}{c}{\textbf{Formula Conservation (\%)}}\\
    \cmidrule(lr){2-3} \cmidrule(lr){4-5} \cmidrule(lr){6-7} \cmidrule(lr){8-9} \cmidrule(lr){10-11}
    & \textbf{Agent} & \textbf{Baseline} & \textbf{Agent} & \textbf{Baseline} & \textbf{Agent} & \textbf{Baseline} & \textbf{Agent} & \textbf{Baseline} & \textbf{Agent} & \textbf{Baseline}\\
    \midrule
    claude-haiku-4.5       & 11.51 &  9.62 & 88.68 & 86.04 & 41.30 & 40.70 & 43.19 & 36.41 & 29.15 & 23.03\\
    claude-opus-4.5        & 25.66 & 28.49 & \textbf{97.74} & \textbf{97.17} & 58.71 & 62.05 & 24.92 & \textbf{15.43} & 61.58 & 56.31\\
    claude-sonnet-4.5      & 18.11 & 17.55 & 96.60 & 95.09 & 50.04 & 51.98 & 40.47 & 34.94 & 49.02 & 42.26\\
    deepseek-v3.1          &  7.36 &  6.79 & 84.34 & 86.23 & 37.31 & 35.42 & 51.97 & 55.58 & 23.71 & 16.41\\
    deepseek-v3.2          & 11.32 &  5.66 & 85.28 & 86.04 & 41.56 & 31.65 & 47.68 & 52.50 & 29.87 & 16.01\\
    deepseek-v3.2-thinking & 16.60 & 20.38 & 71.13 & 56.79 & 53.55 & 64.12 & 27.27 & 23.17 & 64.46 & 81.73\\
    gemini-2.5-pro         & 22.08 & 30.19 & 76.98 & 87.74 & 61.13 & 65.07 & 34.61 & 27.68 & 71.81 & 71.61\\
    gemini-3-flash         & \textbf{51.51} & 51.13 & 94.72 & 95.09 & \textbf{77.69} & \textbf{78.06} & 23.89 & 21.94 & 90.84 & 85.52\\
    gemini-3-pro           & 48.30 & \textbf{52.08} & 96.79 & 96.04 & 74.43 & 77.39 & 24.85 & 23.69 & \textbf{93.57} & \textbf{90.57}\\
    gpt-5.2                & 11.70 &  7.36 & 67.55 & 74.91 & 46.57 & 38.97 & \textbf{24.25} & 24.52 & 35.47 & 23.93\\
    kimi-k2-thinking       & 11.32 & 20.57 & 73.02 & 71.89 & 47.17 & 60.20 & 38.19 & 30.08 & 49.10 & 75.07\\
    qwen3-max              & 15.28 &  4.72 & 83.02 & 89.62 & 45.96 & 31.15 & 48.08 & 57.75 & 47.95 & 13.68\\
    \bottomrule
  \end{tabular}}}

  \end{threeparttable}
\end{table*}

\subsection{Overall Performance}

The comparative performance of evaluated LLMs across both Agent and Baseline configurations is summarized in Table~\ref{tab:main-results}. Our analysis reveals several key insights into the current state of ``LLM-as-a-Chemist.''

\subsubsection{Performance Hierarchy and Foundation Capabilities}
Models exhibit a stark tri-modal distribution in performance. The Frontier Group, led by \textit{Gemini 3 Flash} (51.51\%) and \textit{Gemini 3 Pro} (48.30\%), achieves a significant lead, effectively setting the state-of-the-art for the MolQuest benchmark. Notably, their high baseline accuracy suggests that the \textit{Gemini 3} family possesses a superior internal representation of the spectra-to-structure mapping, likely due to enhanced multi-modal pre-training on aligned chemical data. 

The Mid-tier Group (e.g., \textit{Claude Opus 4.5}, \textit{Gemini 2.5 Pro}) stabilizes between 20--30\% accuracy, while the Struggling Group (e.g., \textit{DeepSeek V3.1}, \textit{Qwen3 Max} in baseline) fails to cross the 10\% threshold. This suggests that for complex molecular elucidation, general-purpose reasoning is insufficient without a robust foundational understanding of chemical topology and spectral interpretation.

\subsubsection{Chemical Consistency and Hallucination Control}
The \textit{Formula Conservation} and \textit{SMILES Validity} metrics provide a rigorous check against stochastic ``guessing.''

High-Fidelity Reasoning: \textit{Gemini 3 Pro} achieves a remarkable 93.57\% formula conservation, indicating a strict adherence to mass-balance constraints derived from MS data. This suggests the model does not merely generate ``plausible-looking'' SMILES but actively constrains its structural search space within the provided molecular formula.
Connectivity Hallucination: Conversely, models like \textit{DeepSeek v3.1} exhibit a disconnect between evidence and generation, with conservation rates as low as 23.71\%. This indicates a ``hallucination-driven'' failure mode where the model ignores spectral constraints to output familiar but incorrect structural motifs.

\subsubsection{Probabilistic Reliability and Self-Assessment}
The \textit{Calibration Error} reveals the models' metacognitive ability---knowing when they are likely to be wrong. \textit{Claude Opus 4.5}, despite not being the most accurate, demonstrates the highest reliability with the lowest calibration error (15.43\% in Baseline). This trait is critical for autonomous research; a well-calibrated model can signal for human intervention when its confidence is low. In contrast, the \textit{Thinking} models (\textit{DeepSeek V3.2 Thinking} and \textit{Kimi K2}) show relatively high calibration errors in the agentic setting, suggesting that internal ``thought traces'' do not always translate into accurate self-assessment of the final structural output.

\subsection{Impact of the Dynamic Paradigm}

Comparing the \textbf{Agent} (dynamic) and \textbf{Baseline} (static) columns in Table~\ref{tab:main-results} reveals a critical bifurcation in how models adapt to interactive environments:

The ``Empowered'' Group: Models such as \textit{Qwen3 Max} (+10.56\%), \textit{DeepSeek v3.2} (+5.66\%), and \textit{GPT-5.2} (+4.34\%) show significant accuracy gains in the Agent mode. For these models, the dynamic framework acts as a cognitive scaffold. By breaking the monolithic elucidation task into sequential steps (e.g., ``Get Formula'' $\rightarrow$ ``Get NMR'' $\rightarrow$ ``Reason''), the agentic workflow reduces working memory load and activates more effective chain-of-thought reasoning.

The ``Challenged'' Group: Conversely, models like \textit{Kimi K2 Thinking} (-9.25\%) and \textit{Gemini 2.5 Pro} (-8.11\%) perform worse in the Agent mode. Interaction logs suggest a deficit in strategic planning: these models struggle to evaluate the ``value of information,'' often making redundant requests or failing to synthesize sequentially acquired evidence as effectively as they handle static context.

This divergence confirms that MolQuest diagnoses not just chemical knowledge, but the intrinsic capability for \textit{strategic planning} under resource constraints.

\begin{figure}[H]
  \centering
  \includegraphics[width=\linewidth]{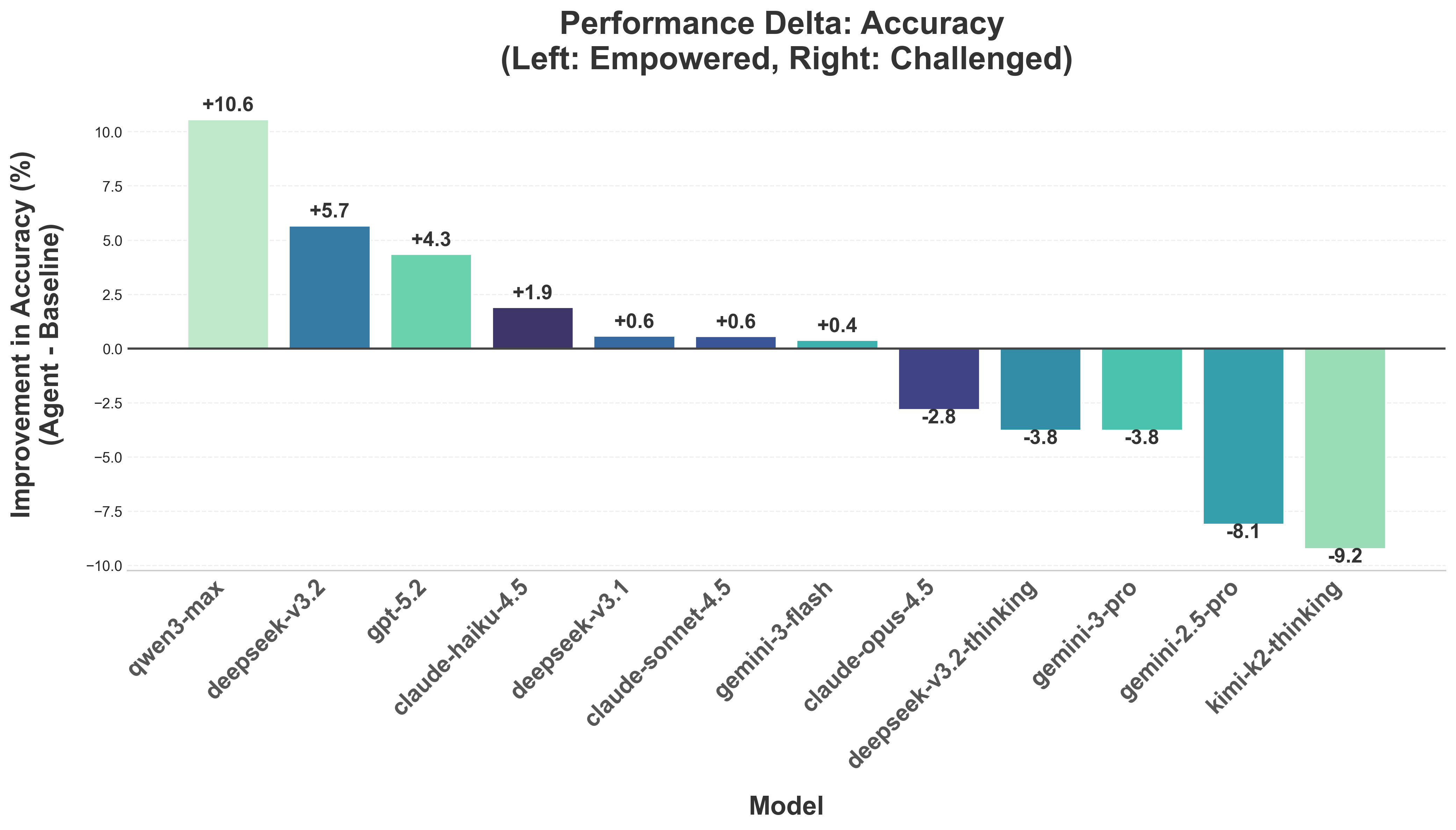}
  \Description{Additional experimental results on MolQuest, illustrating performance differences across models or settings.}

  \caption{the core results of the ablation study (dynamic vs. static)}
  \label{fig:result2}
\end{figure}

\subsection{Interaction Efficiency and Cost-Effectiveness}

To further investigate the behavioral patterns of different models within the agentic framework, we analyze their operational efficiency. We introduce the Average Interaction Rounds (Avg. Rounds) to measure the decisiveness of an agent and Accuracy per 1M Tokens (Acc/1M) as a metric for economic efficiency in complex reasoning.

\begin{table}[b]
  \centering
  \caption{Analysis of interaction efficiency and cost-effectiveness across models.}
  \label{tab:efficiency-results}
  \small
  \setlength{\tabcolsep}{3pt}
  \resizebox{\linewidth}{!}{%
  \begin{tabular}{lccc}
    \toprule
    \textbf{Model} & \textbf{Avg. Rounds} & \textbf{Accuracy (\%)} & \textbf{Acc/1M Tokens} \\
    \midrule
    claude-haiku-4.5       & 4.36 & 11.51 & 4.57 \\
    claude-opus-4.5        & 4.72 & 25.66 & \textbf{9.18} \\
    claude-sonnet-4.5      & 5.02 & 18.11 & 5.88 \\
    deepseek-v3.1          & 5.90 & 7.36  & 2.51 \\
    deepseek-v3.2          & 4.42 & 11.32 & 2.37 \\
    deepseek-v3.2-thinking & 4.61 & 16.60 & 4.78 \\
    gemini-2.5-pro         & 4.51 & 22.08 & 6.11 \\
    gemini-3-flash         & 4.70 & \textbf{51.51} & 8.87 \\
    gemini-3-pro           & 4.81 & 48.30 & 8.84 \\
    gpt-5.2                & \textbf{3.76} & 11.70 & 8.49 \\
    kimi-k2-thinking       & 3.93 & 11.32 & 5.02 \\
    qwen3-max              & 4.99 & 15.28 & 4.44 \\
    \bottomrule
  \end{tabular}
  }
\end{table}

The Pareto Frontier of Intelligence.
As illustrated in Table~\ref{tab:efficiency-results}, \textit{Gemini 3 Flash} and \textit{Gemini 3 Pro} achieve a near-optimal balance between interaction depth and success rate. Both models maintain a moderate interaction frequency ($\sim$4.7-4.8 rounds), suggesting they do not rely on exhaustive data retrieval but rather on targeted information acquisition. Notably, \textit{Claude Opus 4.5} leads in economic efficiency (9.18 Acc/1M Tokens), indicating that while its absolute accuracy is lower than the Gemini 3 series, its reasoning process is highly concise.

Decisiveness vs. Hesitation.
A significant correlation is observed between low accuracy and high interaction rounds in the case of \textit{DeepSeek v3.1} (5.90 rounds). This "interaction trap" suggests that weaker models struggle with terminating logic; they continue to request redundant spectral data without being able to synthesize a coherent structure, leading to a waste of computational resources. Conversely, \textit{GPT-5.2} and \textit{Kimi K2 Thinking} exhibit the lowest average rounds ($<$4.0), which in their case reflects premature termination---submitting incorrect structures before acquiring sufficient NMR evidence.

\section{Discussion}

The experimental results from MolQuest provide a diagnostic assessment of large language models (LLMs) in dynamic scientific reasoning. Our analysis reveals a key distinction: the interactive framework serves as a cognitive scaffold that enhances some models (e.g., Qwen3 Max), while acting as a strategic crucible that exposes planning deficits in others (e.g., Kimi K2 Thinking). This indicates that static benchmarks, which provide all information at once, may conflate pattern recognition with true problem-solving agency. Furthermore, metrics such as Formula Conservation and calibration error highlight critical dimensions of reliability beyond mere accuracy. Many models exhibit "connectivity hallucination," generating chemically plausible structures that violate experimental constraints, whereas well-calibrated models demonstrate essential self-awareness regarding their own uncertainty.

\subsection{Limitations}

This study has several inherent limitations. First, the conclusions are based on a specific domain—small organic molecule elucidation—and a selected set of general-purpose LLMs. Model performance may differ for other scientific tasks (e.g., synthesis planning) or when using domain-specialized models. Second, while the MolQuest framework introduces dynamic interaction, the evaluation remains largely outcome-based rather than process-oriented. The assessment of reasoning quality, such as the logical soundness of hypothesis generation or the optimality of decision sequences, still relies on final accuracy metrics and lacks fine-grained, automated analysis of the reasoning chain itself. Third, the simulated "cost" within the environment is abstract and does not fully capture the real-world economic, temporal, and material constraints of a physical laboratory. Finally, although diverse, the 530-molecule dataset represents a limited segment of chemical space (150-500 Da). Future work should expand the benchmark to include more complex biomolecules and challenging "corner cases" to thoroughly stress-test model robustness and generalizability.

\subsection{Future Research Directions}

The findings from this work point to several important avenues for future research. First, there is a clear need to move from passive capability evaluation to the active design of adaptive interaction protocols that can optimize model performance. This involves creating tailored scaffolds for "empowered" models to maximize efficiency, and developing pedagogically-structured protocols to train strategic planning in "challenged" models. Second, applying the core MolQuest paradigm—interactive, evidence-driven, and resource-constrained problem-solving—to other scientific domains (e.g., materials characterization, genomics) will test the generality of the observed "scaffold versus crucible" effect and help distinguish domain-specific knowledge gaps from fundamental limitations in reasoning. Third, the diagnostic capability profiles generated by MolQuest can directly inform the design of hybrid human-AI collaborative systems. Research should focus on defining optimal collaboration patterns where models and experts leverage complementary strengths, such as using LLMs for rapid hypothesis generation and data triage, while reserving human expertise for high-level strategy and complex validation. In summary, by shifting focus from what models know to how they reason under constraints, this work establishes a foundation for developing more reliable, strategically competent, and collaborative AI systems in science.

\section{Conclusion}

To address the critical gap in evaluating dynamic and strategic reasoning within scientific AI, we introduce \textbf{MolQuest}, a novel benchmark that reframes molecular structure elucidation as an interactive, sequential decision-making task under resource constraints. MolQuest's core innovation lies in its integration of authentic, multi-modal experimental data—including raw NMR and MS spectra—into an agent-based simulation environment. This approach moves beyond synthetic or simplified examples, anchoring tasks directly in real-world chemistry literature and preserving the inherent noise, ambiguity, and information gaps of actual research. Consequently, MolQuest establishes a new evaluation paradigm for the “LLM-as-a-Chemist” that directly mirrors the cognitive and operational challenges of laboratory work.Our comprehensive evaluation yields a core finding: the dynamic interactive framework, characterized by information asymmetry and a "Plan--Request--Reason" loop, does not affect all models uniformly. It functions as a diagnostic lens, acting as a cognitive scaffold that significantly enhances the performance of models capable of strategic information-seeking and abductive reasoning . Conversely, it reveals fundamental deficits in strategic planning in models that falter under these conditions . This bifurcation underscores that future assessment of scientific AI must evolve beyond static, single-turn benchmarks toward diagnostic systems capable of elucidating a model's intrinsic reasoning and planning capabilities.Furthermore, this work demonstrates that well-structured interaction protocols are not merely passive evaluation tools but can serve as active instruments for eliciting and enhancing AI's scientific problem-solving abilities. By providing a concrete framework grounded in real data, a high-quality dataset, and a reproducible methodology, MolQuest lays a foundation for subsequent research aimed at diagnosing AI capabilities and building reliable human-AI collaborative partnerships, thereby accelerating the development of AI-augmented scientific discovery.

\section{Code and Data Availability Statement}
To support reproducibility, the implementation of the MolQuest framework and the associated benchmark dataset will be updated on https://github.com/SKYLENAGE-AI in the near future.

\let\balance\relax
\bibliographystyle{unsrtnat}
\bibliography{references}

\appendix
\clearpage

\subsection*{Appendix Contents}

\noindent A. Benchmark Dataset Details\\
\hspace*{2em}A.1 Data Source and Curation Criteria\\
\hspace*{2em}A.2 Dataset Distribution, and Diversity Analysis\\
\hspace*{4em}A.2.1 Chemical Space Diversity Visualization\\
\hspace*{4em}A.2.2 A Subset of Representative Molecules\\
\hspace*{2em}A.3 Example of a Data Record\\

\noindent B. Technical Details of the Human-in-the-Loop Data Pipeline\\
\hspace*{2em}B.1 Multi-LLMs design\\
\hspace*{2em}B.2 Chemical Validation and Cross-Verification Process\\
\hspace*{2em}B.3 Human Review Interface\\
\hspace*{2em}B.4 Common LLM Failure Modes Requiring Human Intervention\\

\noindent C. Complete Specification of the Agent Interaction Framework\\
\hspace*{2em}C.1 Environment State Machine Definition\\
\hspace*{2em}C.2 Tool List and API Specification\\
\hspace*{2em}C.3 Core Agent Prompt\\
\hspace*{2em}C.4 Ablation Study Control: Static One-Shot Input Mode\\

\noindent D. Additional Experimental Results and Analysis\\
\hspace*{2em}D.1 Complete Main Results Table\\
\hspace*{2em}D.2 Ablation Studies under Different Configurations\\

\noindent E. Code and Data Availability Statement\\
\hspace*{2em}E.1 Code Repository\\
\hspace*{2em}E.2 Benchmark Data Access

\section{Benchmark Dataset Details}

\subsection{Data Source and Curation Criteria}
The benchmark dataset was constructed by extracting chemical data from high-impact organic chemistry journals, specifically targeting the timeframe of \textbf{2025--2026}. Primary sources include:
\begin{itemize}
    \item \textit{Journal of American Chemical Society}
    \item \textit{JACS Au}
    \item \textit{Chemistry --- A European Journal}
    \item \textit{Chemical Science}
    \item \textit{ACS Sustainable Chemistry \& Engineering}
    \item \textit{Journal of Physical Organic Chemistry}
    \item \textit{Nature}
    \item \textit{Nature Communications}

\end{itemize}

\textbf{Curation Criteria:}
\begin{enumerate}
    \item \textbf{Data Completeness:} Compounds must contain both independent $^1$H NMR and $^{13}$C NMR spectroscopic data.
    \item \textbf{Novelty:} Only newly synthesized compounds (as defined in the original literature) were included.
    \item \textbf{Exclusion Rules:} Organometallic complexes and polymers were excluded to focus on small-molecule organic structural elucidation.
\end{enumerate}

\subsection{Dataset Distribution, and Diversity Analysis}

\subsubsection{Chemical Space Diversity Visualization}
The chemical space of the benchmark set was analyzed using molecular fingerprints and dimensionality reduction.

\begin{figure}[H]
    \centering
    \includegraphics[width=0.8\linewidth]{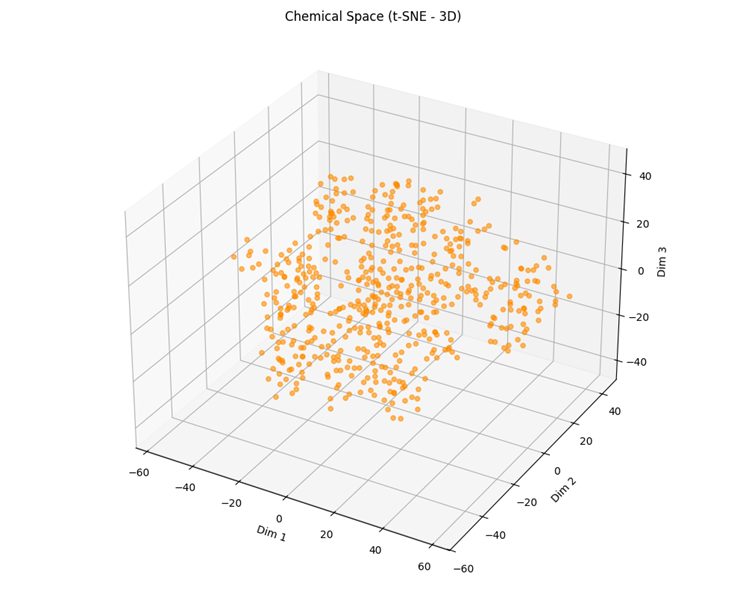}
    \Description{Chemical space map of the benchmark compounds based on molecular fingerprints and t-SNE.}
    \caption{Chemical space visualization of the benchmark molecular set based on molecular fingerprints and t-SNE dimensionality reduction.}
    \label{fig:S2}
\end{figure}

\begin{figure}[H]
    \centering
    \includegraphics[width=0.8\linewidth]{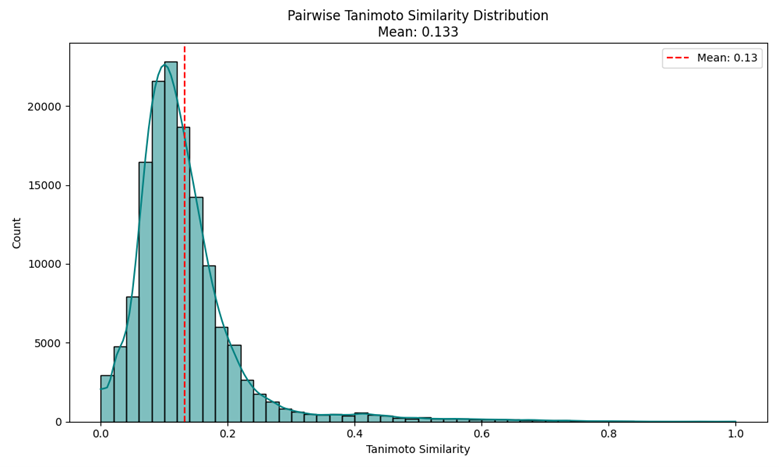}
    \Description{Histogram or density plot showing the distribution of pairwise Tanimoto similarities among benchmark molecules.}
    \caption{Statistics of pairwise Tanimoto similarity distribution for the benchmark molecular set.}
    \label{fig:S3}
\end{figure}

\begin{figure}[H]
    \centering
    \includegraphics[width=0.8\linewidth]{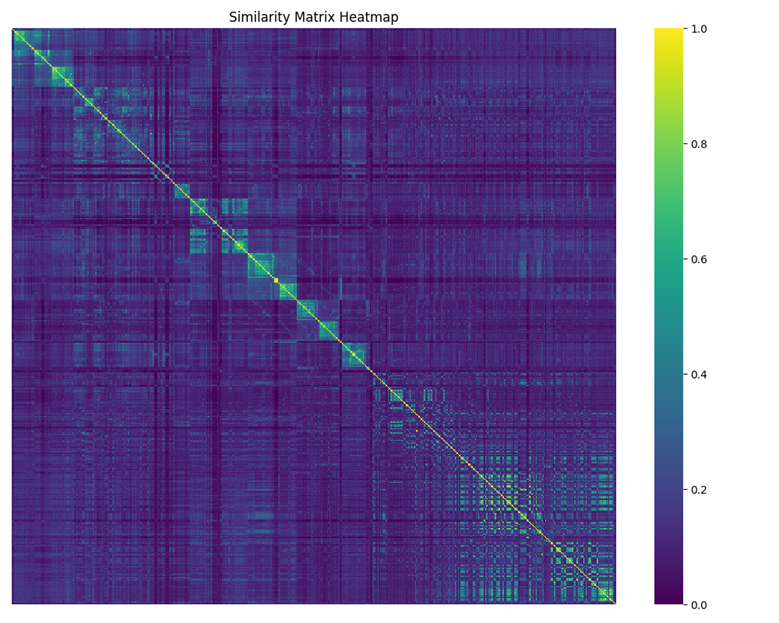}
    \Description{Heatmap visualization of pairwise molecular similarity for the benchmark set based on molecular fingerprints.}
    \caption{Visualization of the pairwise similarity matrix heatmap of the benchmark molecular set based on molecular fingerprints.}
    \label{fig:S4}
\end{figure}

\subsubsection{A Subset of Representative Molecules}
[Detailed description of representative structural motifs present in the dataset.]

\subsection{Example of a Data Record}
Below is a structured JSON mapping for a de-identified molecule, demonstrating the transformation from raw text to our machine-readable format.

\begin{lstlisting}[language=json, caption=Example of a structured data record.]
{
  "molecule_name": "Dineopentyl sulfite",
  "uuid": "a6dbee66-c2af-45ed-92d4-cab27372530b",
  "smiles": "CC(C)(C)CO[S@@](=O)OCC(C)(C)C",
  "molecular_formula": "C10H22O3S",
  "molecular_weight": 222.35,
  "inchi": "",
  "inchi_key": "",
  "raw_data": {
    "1H_NMR": "1H NMR (300 MHz, CDCl3): \\delta 3.70 (d, J = 9.6 Hz, 2H), 3.53 (d, J = 9.6 Hz, 2H), 0.95 (s, 18H).",
    "13C_NMR": "13C NMR (75 MHz, CDCl3): \\delta 71.24, 31.65, 26.34.",
    "19F_NMR": null,
    "31P_NMR": null,
    "IR_film": null,
    "IR_neat": null,
    "HRMS_ESI": "HRMS (ESI-TOF) m/z: [M+H]+ Calcd for C10H22SO3: 223.1368 g/mol",
    "MS_EI": null,
    "HRMS_EI": null,
    "MS_APCI": null,
    "HRMS_APCI": null,
    "HRMS_CI": null,
    "Melting_Point": null,
    "TLC": null,
    "Optical_Rotation": null
  }
}
\end{lstlisting}

\section{Technical Details of the Human-in-the-Loop Data Pipeline}

\subsection{Multi-LLMs design}
The pipeline utilizes three core:
\begin{itemize}
    \item \textbf{Segmenter:} Responsible for isolating specific spectral data from raw PDF/text.
    \item \textbf{Spectroscopist:} Extracts and structures data into JSON (shifts, multiplicities, integrals).
    \item \textbf{Judge:} Performs logical checks and consistency verification.
\end{itemize}

\subsection{Chemical Validation and Cross-Verification Process}
\begin{itemize}
    \item \textbf{External APIs:} We utilized \texttt{PubChemPy} and \texttt{OPSIN} for SMILES-to-name and name-to-structure cross-checks.
    \item \textbf{LLM Cross-Verification:} Specific prompts were designed to check internal consistency, such as comparing the sum of NMR integrals against the hydrogen count in the molecular formula.
\end{itemize}

\subsection{Human Review Interface}
We developed a Streamlit-based interface to facilitate the "Human-in-the-Loop" stage.
\begin{figure}[H]
    \centering
    \includegraphics[width=0.9\linewidth]{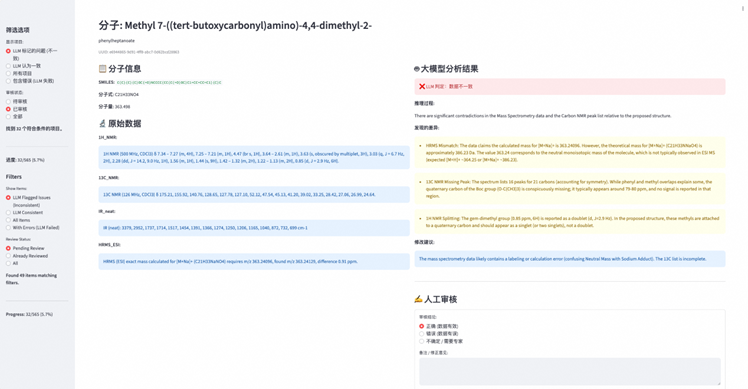}
    \Description{Screenshot of the human-in-the-loop review interface, highlighting flagged entries and the red-flag review workflow.}
    \caption{Screenshot of the human review interface showing the "Red Flag" system for flagged data points.}
\end{figure}

\subsection{Common LLM Failure Modes Requiring Human Intervention}
During the data extraction process, the following systematic errors were identified in LLM outputs:
\begin{enumerate}[label=\alph*.]
\raggedright
    \item \textbf{Exchangeable Protons:} Frequent misinterpretation of missing hydroxyl ($-OH$) or carboxylic acid ($-COOH$) protons in NMR data.
    \item \textbf{SMILES Atom Counting:} Enumeration errors when calculating atom counts from SMILES strings.
    \item \textbf{Peak Deviation:} Incorrect flagging of valid peaks due to minor chemical shift deviations.
    \item \textbf{MS Adducts:} Confusion between calculated molecular mass and observed mass spectrometry peaks (e.g., $[M+H]^+$ or $[M+Na]^+$ vs. $M$).
\end{enumerate}

---

\section{Complete Specification of the Agent Interaction Framework}

\subsection{Environment State Machine Definition}
The agent interaction is modeled as a state machine where:
\begin{itemize}
    \item $s_t = \{messages, known\_data, costs\}$ represents the state at time $t$.
    \item $T(s_t, a_t) \rightarrow s_{t+1}$ is the transition function, where $a_t$ is the action (tool call) taken by the agent.
\end{itemize}

\subsection{Tool List and API Specification}

\subsection{Core Agent Prompt}
The following system prompt defines the "Expert Spectroscopist" persona:

\begin{lstlisting}[caption={Initial Prompt (Dynamic Multi-step Mode)},basicstyle=\ttfamily\footnotesize,breaklines=true]
You are a senior expert in organic synthesis and spectroscopic structure elucidation,
working in a laboratory to analyze an unknown small organic molecule with ID {sample_id}.
You may call tools to obtain MS, 1H NMR, 13C NMR, and other data, and then infer its structure.

[Your Objectives]
1. Use all available spectroscopic and molecular information to elucidate the structure.
2. When information is insufficient, plan and call only the necessary tools to collect evidence.
3. Provide the most plausible candidate structure (as SMILES) and a confidence score.
4. Efficiency matters (important): each tool call simulates a real experiment with time and monetary cost.
   Prioritize correctness of the final structure, but minimize the number of tool calls by reasoning deeply
   from the data already available.

[Available Tools]
The tool list below is injected at runtime and always matches the tools you can actually call:

{tools_description}

[System Behavior Note (important)]
- To avoid invalid calls: if you do not call any tool at the very beginning, the system may automatically
  run Check_Data once to confirm which data are available for the current sample. Unless you truly need to
  verify availability, you do not have to treat Check_Data as a mandatory first step.

[Suggested Reasoning Workflow (adjust as needed)]
1. Strategy planning: identify the most critical uncertainty first; avoid blindly requesting all spectra.
   If you are unsure whether certain data exist or want to avoid wasted calls, you may call Check_Data.
2. Iterative data acquisition (important): acquire data step-by-step. After each tool result, pause to
   reason and decide whether you are still uncertain. Only then request the next most informative test.
   Prefer calling at most ONE new tool per iteration unless multiple calls are strictly necessary.
3. Basic data: call Measure_MW and Measure_Formula to obtain molecular weight and molecular formula.
4. First pass: use Calculate_DBE to estimate unsaturation and form initial hypotheses.
5. Core elucidation: prioritize Get_1H_NMR (and Get_13C_NMR only if needed). Use chemical shifts,
   integrals, splitting patterns, and carbon environments to build one or more plausible scaffolds.
6. Targeted validation: only when ambiguity remains, cautiously call Get_IR / Get_HRMS (etc.) to obtain
   specific discriminating evidence.
7. Once you have the most plausible candidate structure, provide its SMILES.
8. If the evidence is still clearly insufficient for a unique structure, explicitly state what is uncertain,
   and provide your best current candidate with an appropriate confidence.

[Output Requirements (very important)]
When you believe you are done and no further tool calls are needed, your final reply must include the
following structured block (key names must match exactly). You may provide natural-language reasoning
before it, but the block below is mandatory:

FINAL_RESULT:
  UUID: {sample_id}
  PREDICTED_SMILES: <SMILES of your best candidate; if you cannot give one, write "UNKNOWN">
  CONFIDENCE: <a decimal between 0 and 1, e.g., 0.8>
  REASON_BRIEF: <1--3 sentences summarizing the key evidence supporting your choice>

Notes:
- If multiple candidates are plausible, you must output only your top choice in PREDICTED_SMILES;
  you may mention other candidates in the free-text reasoning.
- If you believe the data are insufficient for a reliable structure, you must set PREDICTED_SMILES="UNKNOWN"
  and explain why in REASON_BRIEF.

The unique identifier of the current sample is {sample_id}. Start by planning which tools to call,
then proceed step by step to complete the structure elucidation.
\end{lstlisting}

\subsection{Ablation Study Control: Static One-Shot Input Mode}
To evaluate the benefit of the dynamic interaction, we defined a Static One-Shot Baseline.

\textbf{a. System Prompt for Static Mode:}
\begin{lstlisting}[caption={System Prompt (Static Mode)},basicstyle=\ttfamily\footnotesize,breaklines=true]
You are a senior expert in organic synthesis and spectroscopic structure elucidation.
All available raw data for sample {sample_id} are provided upfront.
Use only the provided data and do not call any tools.
When finished, output the required FINAL_RESULT block exactly in the specified format.
\end{lstlisting}

\textbf{b. Input data format (Static Mode).}
In static mode, all relevant experimental data (including molecular formulas, molecular weights, and raw data from various types of spectra) are consolidated into a JSON object and injected into the user prompt as the \texttt{\{raw\_json\}} parameter in a single operation.

\begin{lstlisting}[language=json,caption={Example integrated JSON data (Static Mode)},basicstyle=\ttfamily\footnotesize,breaklines=true]
{
  "uuid": "337aeb0a-8c68-44c0-b891-444a9b6a9c1d",
  "molecular_formula": "C9H8O3S",
  "molecular_weight": 196.227,
  "raw_data": {
    "1H_NMR": "1H NMR (400 MHz, CDCl3): \\delta 7.62--7.66 (m, 2H), 7.57 (tt, J = 7.5, 2.1 Hz, 1H), \\delta 7.43--7.48 (m, 2H), \\delta 4.09 (s, 3H)",
    "13C_NMR": "13C NMR (100 MHz, CDCl3): \\delta 133.01, 131.96, 128.88, 117.24, 91.32, 78.87, 58.07",
    "19F_NMR": null,
    "31P_NMR": null,
    "HRMS_ESI": "HRMS (ESI+): m/z calc'd for C9H12NO3S [M+NH4]+: 214.0532, found: 214.0531",
    "IR_film": null,
    "Melting_Point": null
    // ... other empty fields omitted for brevity
  }
}
\end{lstlisting}

\textbf{c. Task Instructions \& Output Requirements.}
The user prompt combines the sample ID, the aforementioned JSON data, and strict output-format instructions. The model is required to directly output the prediction upon receiving the data, without intermediate interaction.

\begin{lstlisting}[caption={User Prompt (Static Mode)},basicstyle=\ttfamily\footnotesize,breaklines=true]
Sample UUID: {sample_id}
Raw data (JSON, provided as-is):
{raw_json}

Output format (must match exactly):
FINAL_RESULT:
  UUID: {sample_id}
  PREDICTED_SMILES: <SMILES of your best candidate; if you cannot give one, write "UNKNOWN">
  CONFIDENCE: <a decimal between 0 and 1, e.g., 0.8>
  REASON_BRIEF: <1--3 sentences summarizing the key evidence supporting your choice>

Explanation:
  {sample_id}: The unique identifier (UUID) of the sample.
  {raw_json}: The JSON string containing complete experimental data as described above.
  FINAL_RESULT: To ensure comparability of evaluation metrics (e.g., accuracy and SMILES matching), static mode mandates a structured output block identical to that used in dynamic mode.
\end{lstlisting}

\section{Additional Experimental Results and Analysis}

\subsection{Complete Main Results Table}
[Insert comprehensive table here comparing all 13 models across Accuracy, F1, Tanimoto, and Cost.]

\subsection{Ablation Studies under Different Configurations}
Analysis of how limiting the maximum interaction rounds ($N_{max}$) impacts the final confidence scores and accuracy.

\section{Code and Data Availability Statement}

\subsection{Code Repository}
The complete source code for the multi-agent framework, including modules for \texttt{features/} and \texttt{verify\_data/}, is available in an anonymized repository (link omitted for double-blind review).

\noindent\textbf{Repository link placeholder:} \url{REPOSITORY_URL_PLACEHOLDER}

\subsection{Benchmark Data Access}
The structured benchmark dataset is archived on Zenodo (DOI: [Insert DOI]). To protect journal copyright, we provide structured JSON records derived from the text rather than original PDF files.

\noindent\textbf{Zenodo placeholder:} DOI: \texttt{ZENODO\_DOI\_PLACEHOLDER} \quad (or \url{ZENODO_URL_PLACEHOLDER})

\end{document}